%% file: 0_main.tex
\documentclass{article}

\usepackage[preprint,nonatbib]{neurips_2022}

\usepackage[utf8]{inputenc} 
\usepackage[T1]{fontenc}    
\usepackage{hyperref}       
\hypersetup{
    colorlinks=true,
    linkcolor=blue,
    filecolor=magenta,      
    urlcolor=cyan,
}
\usepackage{url}            
\usepackage{booktabs}       
\usepackage{amsfonts}       
\usepackage{nicefrac}       
\usepackage{microtype}      
\usepackage{xcolor,colortbl}         

\usepackage{enumitem,kantlipsum}
\usepackage{multirow}
\usepackage{graphicx}
\usepackage{amsmath}
\usepackage{cleveref}
\usepackage{wrapfig}
\usepackage{lipsum}
\usepackage[export]{adjustbox}
\usepackage{caption}

\newcommand{\circlednumber}[1]{\textcircled{\raisebox{-0.9pt}{#1}}}
\newcommand{\overnumber}[2]{\stackrel{\textcircled{\raisebox{-0.9pt}{#1}}}{#2}}

\newcommand{\etal}{\textit{et. al.}}
\newcommand{\eg}{\textit{e.g.}}
\newcommand{\ie}{\textit{i.e.}}
\newcommand{\wrt}{\textit{w.r.t.}}
\newcommand{\Ours}{Hessian and Eigen-decomposition-based Adversarial Detection}
\newcommand{\ours}{HEAD}
\newcommand{\HessianFeat}{Hessian Feature}
\newcommand{\hessianfeat}{HF}

\newcommand{\PCAComponent}{Least Significant Component}
\newcommand{\PCAFeat}{Least Significant Component Feature}
\newcommand{\pcafeat}{LSCF}

\newcommand{\expectation}{\mathbf{E}}
\newcommand{\variance}{\mathrm{Var}}

\title{Attack-Agnostic Adversarial Detection}

\author{%
  Jiaxin Cheng \quad Mohamed Hussein  \quad Jay Billa \quad Wael AbdAlmageed \\
  University of Southern California, Information Sciences Institution \\
  \texttt{{chengjia,mehussein,jbilla,wamageed}@isi.edu} \\
}

\begin{document}

\maketitle

\begin{abstract}
  The growing number of adversarial attacks in recent years gives attackers an advantage over defenders, as defenders must train detectors after knowing the types of attacks, and many models need to be maintained to ensure good performance in detecting any upcoming attacks. We propose a way to end the tug-of-war between attackers and defenders by treating adversarial attack detection as an anomaly detection problem so that the detector is agnostic to the attack. We quantify the statistical deviation caused by adversarial perturbations in two aspects. The \PCAFeat\ (\pcafeat) quantifies the deviation of adversarial examples from the statistics of benign samples and \HessianFeat\ (\hessianfeat) reflects how adversarial examples distort the landscape of models' optima by measuring the local loss curvature. Empirical results show that our method can achieve an overall ROC AUC of 94.9\%, 89.7\%, and 94.6\% on CIFAR10, CIFAR100, and SVHN, respectively, and has comparable performance to adversarial detectors trained with adversarial examples on most of the attacks.
  
\end{abstract}

\input{1_introduction}

\input{2_related_work}

\input{3_adv_detect_as_outlier_detection}
\input{5_experiments}

\input{6_conclusion}

{\small
\bibliographystyle{ieee_fullname}
\bibliography{egbib}
}

\appendix

\clearpage
\input{appendix}

\end{document}

%% file: 1_introduction.tex
\section{Introduction}

Despite the success of deep neural networks (DNNs) in computer vision~\cite{alexnet,vgg,resnet}, natural language processing~\cite{lstm,attention,bert} and speech recognition~\cite{asr1,asr2}, DNNs are notoriously vulnerable to adversarial attacks\cite{advattack} that inject carefully crafted imperceptible perturbations into the input and are able to deceive the model with a great chance of success. 

There are three main orthogonal approaches for combating adversarial attacks --- (i) Using adversarial attacks as a data augmentation mechanism by including adversarially perturbed samples in the training data to induce robustness in the trained model~\cite{AT,TRADES,FAT,GAIRAT,RST,MART,CAT,DAT}; (ii) Preprocessing the input data with a denoising function or deep network \cite{denoise1,denoise2,denoise3,denoise4,denoise5} to counteract the effect of adversarial perturbations; and (iii) Training an auxiliary network to \emph{detect} adversarial examples and deny providing inference on adversarial samples~\cite{metzen2017detecting,li2017adversarial,rouhani2017curtail,rouhani2018towards,grosse2017statistical,safetynet,lid,mahalanobis,nnif,spectraldefense,libre,lng}. Our work falls under adversarial example detection as it does not require retraining the main network (as in adversarial training) nor degrade the input quality (as in preprocessing defenses).


Existing adversarial example detection methods~\cite{safetynet,lid,mahalanobis,nnif,spectraldefense} need to train auxiliary networks in a binary classification manner (\eg\ benign versus adversarial attack(s)). The shortcoming of this strategy is that the detector is trained on specific attack(s) that are available and known at training time. To ensure good detection performance at inference time, the detection network needs to be trained on a large number of attacks. Otherwise, the detection network will perform poorly on unseen attacks during training (\ie\ out of domain) or even on seen attacks during training due to overfitting~\cite{rice2020overfitting}. We argue that a good adversarial detection method should be able to detect any adversarial attack, even if the defender is unaware of the type of adversarial attack. To this end, we propose to frame the adversarial sample detection as an anomaly detection problem, in which only one detection model is constructed and trained on only benign samples, such that the detection model is \emph{attack-agnostic}.


We propose an anomaly detection framework for identifying adversarial examples by measuring the statistical deviations caused by adversarial perturbations. We consider the deviation of two complementary features that reflect the interaction of adversarial perturbation with the data and models. The first feature is \PCAFeat\ (\pcafeat), which maps data to a subspace where the distribution of benign images is compact, while the distribution of adversarial images is spread. The second feature is \HessianFeat\ (\hessianfeat), which uses the second order derivatives as a measure of the distortion caused by adversarial perturbation to the model's loss landscape. Our results underscores the utility of each of the two features and their complementary nature.


The contributions of this paper are:
\begin{enumerate}[leftmargin=*, noitemsep, topsep=0pt]
    \item An anomaly detection framework for adversarial detection that measures statistical deviation caused by adversarial perturbations on two proposed features, \pcafeat\ and \hessianfeat, which are theoretically justified and capture the interaction of adversarial perturbations with data and model. 
    \item Empirical analysis demonstrating the effectiveness of our method on detecting eight different adversarial attacks on three datasets. Our method achieves 94.9\% AUC on CIFAR10, 89.7\% AUC on CIFAR100 and 94.6\%AUC on SVHN.
    \item Comprehensive evaluation showing the computational efficiency, sensitivity to hyper parameters, cross-model generalization, and closeness to binary classification upper bound of the proposed anomaly detection method.
\end{enumerate}

%% file: 2_related_work.tex
\section{Related Work}\label{sec.related_work}

\noindent \textbf{Adversarial Attacks} deceive DNNs by adding carefully crafted perturbations that are imperceptible to humans. Attacks can be classified according to their perturbation constraints. Typical attacks such as Fast Gradient Sign Method (FGSM)~\cite{fgsm}, Basic Iterative Method (BIM)~\cite{bim} and Projected Gradient Method (PGD)~\cite{pgd} are $l_{\infty}$ attacks which allow perturbation of all pixels but limit the maximum deviation a pixel can have. $l_2$ attacks, such as Carlini \& Wagner (CW)~\cite{cw} and DeepFool~\cite{deepfool}, limit the total deviation of all pixels that an example can have. $l_0$ attacks try to change as few pixels as possible but allow more perturbation budget for modified pixels, including one-pixel-attack (OnePixel)~\cite{onepixel} and SparseFool~\cite{sparsefool}. Recently, AutoAttack~\cite{autoattack} can better deceive DNN  models by combining AutoPGD~\cite{autoattack}, Fast Adaptive Boundary Attack (FAB)~\cite{fab} and Square attack~\cite{square}. 

\noindent \textbf{Detecting Adversarial Examples} The general approach for detecting adversarial attacks is to train an auxiliary model using benign and adversarial examples. Various network architectures and features have been used. Metzen \etal~\cite{metzen2017detecting} use the activations to train subnetwork detectors. Li \etal~\cite{li2017adversarial} use PCA projected features  to train cascaded detectors. Statistics of the examples are used in ~\cite{rouhani2017curtail,rouhani2018towards}. Grosse \etal~\cite{grosse2017statistical} use Bayesian Uncertainty and train a logistic regression detector. Lu \etal~\cite{safetynet} train quantized RBF-SVM classifier on top of the penultimate ReLU features. Ma \etal~\cite{lid} use the feature distance of example to its nearest neighbours as images' fingerprint. Lee \etal~\cite{mahalanobis} train classifier with confidence score computed from Mahalanobis distance under Gaussian discriminant analysis. Cohen \etal~\cite{nnif} leverage influence function~\cite{influencefunction} and fit a k-NN model to detect adversarial examples. Harder \etal~\cite{spectraldefense} convert examples into frequency domain and detect adversarial examples using Fourier features. Deng \etal~\cite{libre} detect adversarial examples by converting models into Bayesian neural networks. Abusnaina \etal~\cite{lng} use neighborhood connectivity and graph neural network to detect adversarial examples. These mentioned approaches, while performing well, are all trained with supervision, limiting their generalization to unseen attacks. In contrast, our method is trained in anomaly detection fashion and thus generalizes better to unseen attacks.

\noindent \textbf{Anomaly Detection} aims to detect unusual samples in data. Classic approaches includes One-class SVM~\cite{ocsvm}, Random Forest~\cite{randomforest}, Kernel Density Estimation~\cite{kde}, Local Outlier Factor~\cite{lof}, Elliptic Envelope~\cite{ellipticenvelope}. Deep Anomaly Detection~\cite{dsvdd,salehi2021multiresolution,reiss2021panda,li2021cutpaste,mohseni2020self,beggel2019robust,yi2020patch,pang2019deep,bergmann2020uninformed,deecke2018image,park2020learning,perera2019ocgan,perera2019learning,golan2018deep} takes the advantage of DNNs to have better scalability and performance on high dimensional data. In this work, we apply classic anomaly detection approaches for detecting adversarial examples as we find that deep learning-based anomaly detectors are not adequate for detecting adversarial attacks since they learn image-level semantic representations that cannot capture local image subtleties introduced by adversarial attacks. 




%% file: 3_adv_detect_as_outlier_detection.tex
\section{Attack-Agnostic Adversarial Detection}\label{sec:h3ad}

We present our \Ours\ (\ours) by first motivating adversarial detection as an anomaly detection problem. Then, we introduce \PCAFeat\ and \HessianFeat\ and explain the rationale for using them to detect adversarial attacks.

\subsection{Challenges And Rationale}\label{sec:rationale}

A fundamental assumption of existing adversarial attack detection \cite{metzen2017detecting,li2017adversarial,rouhani2017curtail,rouhani2018towards,grosse2017statistical,safetynet,lid,mahalanobis,nnif,spectraldefense,libre,lng} as well as  adversarial augmentation methods  ~\cite{safetynet,lid,mahalanobis,nnif,spectraldefense} is that adversarial attacks are known and samples can easily be generated using these attacks to train the detector or augment the main model being defended. This assumption, however, is not realistic, since more often than not the defender does not know the attacks \emph{a priori} and therefore samples cannot be easily generated to train a supervised detector or train an adversarially-augmented model.

The absence of attack samples (\ie\ negative) training examples  and the need to be attack-agnostic both motivate framing the adversarial detection as an anomaly detection problem. More formally, the task of the defender is to protect the model trained on \emph{only benign examples} $X$ against adversarial examples $\widehat{X}$ that are unknown during training. The detector $D$ will be trained only on benign examples and will give a score $s(x)=D(f(\mathbf{x}))$ for each testing sample $\mathbf{x}$, indicating the likelihood that $\mathbf{x}$ is an adversarial attack, where  $f \in \mathbb{R}^m \times \mathbb{R}^n$ is a feature extraction function and $m$ and $n$ are  the dimensions of input and feature spaces, respectively. The feature extractor could be any hand-crafted function, such as principal component analysis (PCA), or method crafted specifically for adversarial detection such as LID~\cite{lid} and Mahalanobis~\cite{mahalanobis}. Similarly, $D$ can be any arbitrary anomaly detector, \eg, classic approaches such as kernel density estimator~\cite{kde} and One Class SVM~\cite{ocsvm}, or DNN-based methods like DSVDD~\cite{dsvdd}.



We propose to detect adversarial examples using two complementary features of the image that reflect the interaction between adversarial perturbation and dataset as well as DNN models, respectively. The first property, namely \PCAFeat, quantifies the deviation of adversarial examples from the statistics of benign samples by eigen-decomposing the dataset and mapping the data to the eigenvectors that have \textit{smallest} eigenvalues. The second property, namely \HessianFeat, distinguishes adversarial from benign images by inspecting the curvature of the loss landscape locally at the geometric optima of the model, which can be measured by the Hessian matrix of the loss \wrt\ to the inputs (or intermediate layer outputs).

\subsection{\PCAFeat\ (\pcafeat)}\label{sec.pca}


As mentioned in \Cref{sec.related_work}, we hypothesize that extracting image features that capture global context will not work well, since they tend to miss small subtleties introduced by the adversarial perturbations. Therefore, we hypothesize that we need to extract image features that are sensitive to small imperceptible image noise. To extract the \pcafeat, we use principal component analysis (PCA) to project the raw benign images to a new space with orthonomal basis (\ie\ eigenvectors) in which different dimensions are linearly uncorrelated. Rather than retaining the projections that correspond to the largest eignvalues (\ie\ eigenfaces \cite{139758}), we retain the projections on the directions with smallest eignvalues.  Hence, the features consist of the least significant components of the data.

\begin{figure}[!t]
    \centering
    \begin{minipage}[t]{0.32\textwidth}
    \centering
        \includegraphics[height=0.9\linewidth,trim=3px 7px 5px 3px]{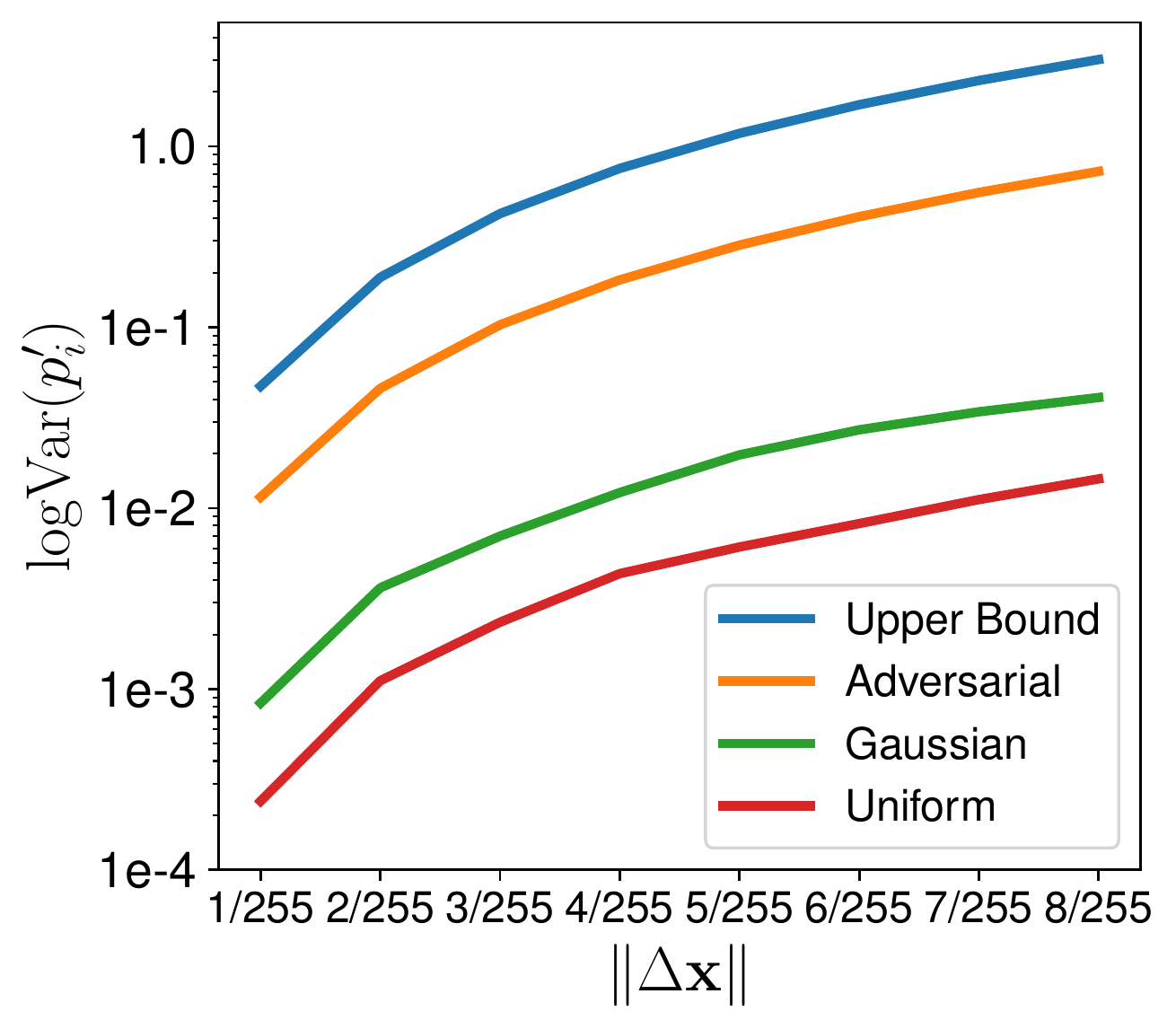}
        \caption{The gap between upperbound in \cref{eq.upperbound_of_derivation} and deviations of $p_i$ caused by adversarial (FGSM), Gaussian and Uniform perturbation on CIFAR10.}
        \label{fig:tightness}
    \end{minipage}\hfill
    \begin{minipage}[t]{0.32\textwidth}
    \centering
        \includegraphics[height=0.9\linewidth,trim=2px 5px 5px 2px]{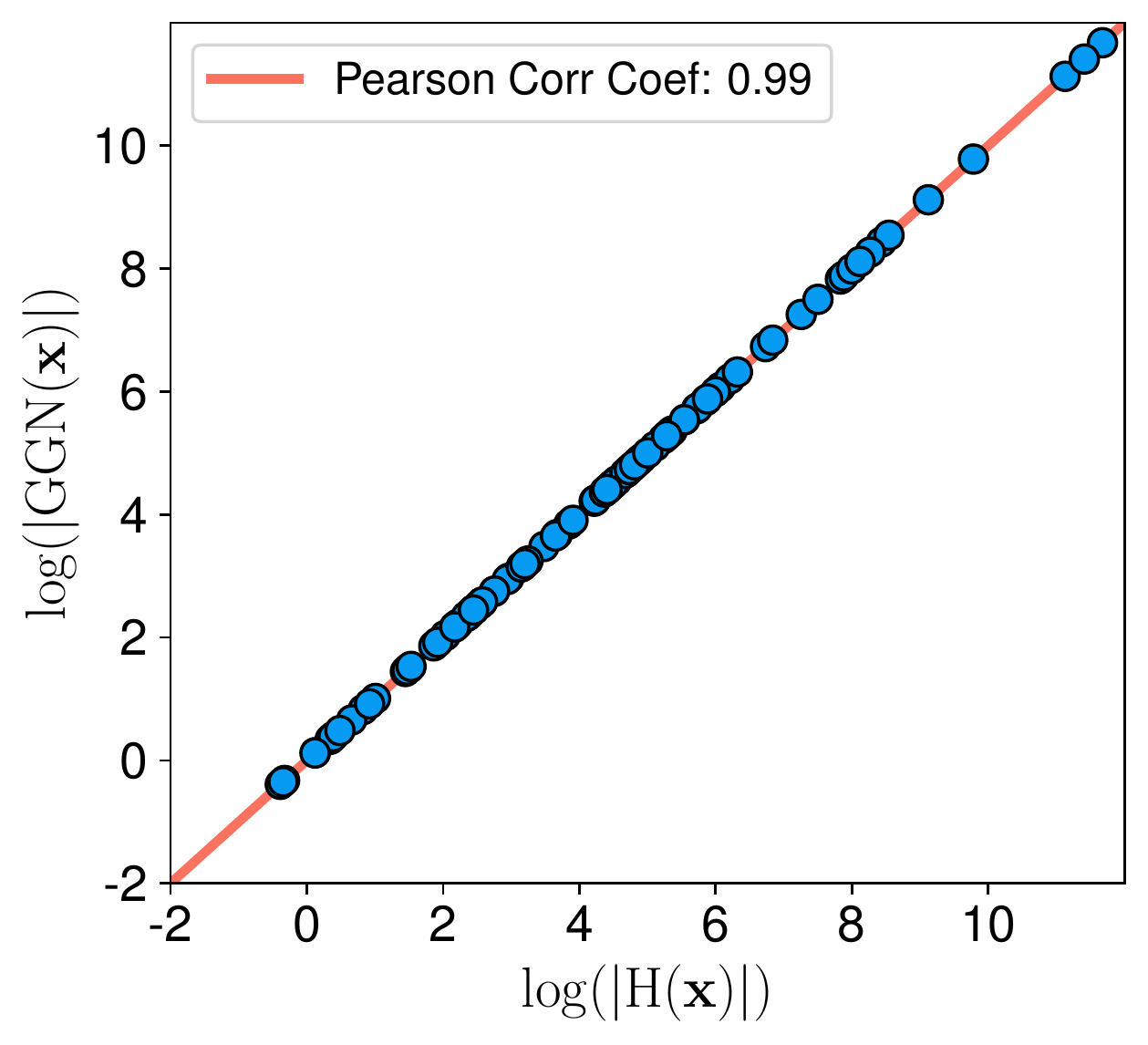}
        \caption{As a good approximation for Hessian, there is a strong correlation between the matrix modulus of GGN and Hessian.}
        \label{fig:GGN_Hessian_approximation}
    \end{minipage}\hfill
    \begin{minipage}[t]{0.32\textwidth}
    \centering
        \includegraphics[height=0.9\linewidth,trim=2px 5px 5px 3px]{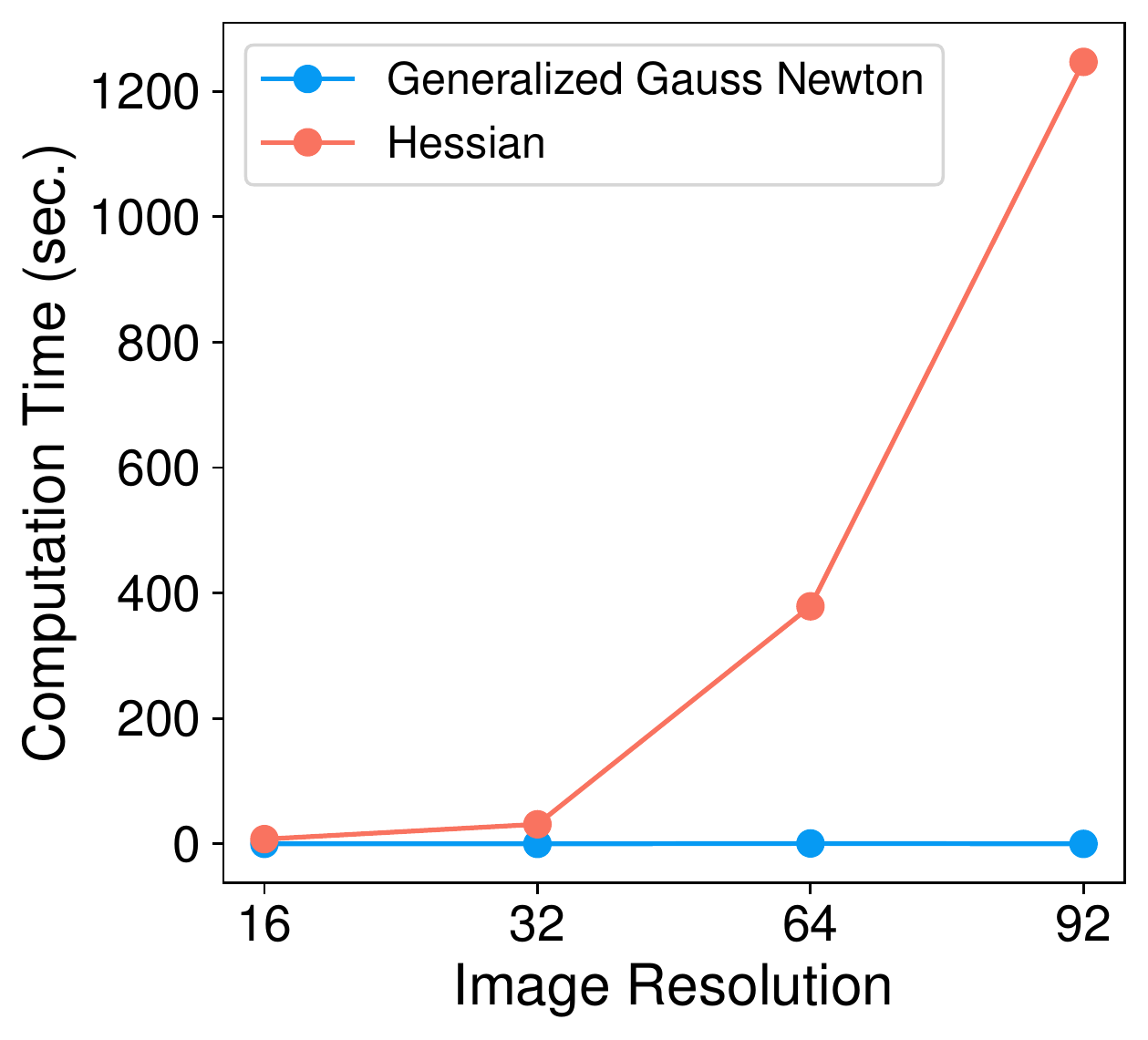}
        \caption{The computation time of Hessian and GGN under different image sizes.}
        \label{fig:GGN_Hessian_time}
    \end{minipage}

\end{figure}

Suppose that the training data $\mathbf{D} \in \mathbb{R}^{N \times m}$ has $N$ samples and $m$ input dimension. Its covariance matrix $\mathbf{C} = \mathbf{D}^{\top} \mathbf{D} / (N-1)$ can be decomposed into $\mathbf{C} = \mathbf{V}\mathbf{L}\mathbf{V}^\top$, where the columns of $\mathbf{V}=[\mathbf{v}_1 \mathbf{v}_2 ... \mathbf{v}_m]$ are the eigenvetors of $\mathbf{C}$, and $\mathbf{L}$ is a diagonal matrix having eigenvalues of $\mathbf{C}$ in descending order on its diagonal (\ie, $\mathrm{diag}(\mathbf{L})=[\lambda_1 \lambda_2 ... \lambda_m], \lambda_i \geq \lambda_{i+1} \forall i \in \{1 .. m-1\}$). The \pcafeat\ of image $\mathbf{x}$ is transformed by the eigenvectors $\mathbf{v}$ that have the smallest eigenvalues. 
\begin{equation}
    f_{\pcafeat}(\mathbf{x}) = \mathbf{x}^\top \mathbf{v} \in \mathbb{R}^{1 \times d} 
\end{equation}
where $d$ is the dimension of \pcafeat\ and $\mathbf{v}=[\mathbf{v}_m \mathbf{v}_{m-1}... \mathbf{v}_{m-d+1}] \in \mathbb{R}^{m \times d}$ is the last $d$ columns of $\mathbf{V}$. We explain the reason for mapping images on the least significant eigenvectors by estimating an upper bound on the expected deviation caused by perturbation for different eigenvectors. Let $p_i=\mathbf{x}^\top \mathbf{v}_i$ be the mapping of image $\mathbf{x}$, and $p'_i=(\mathbf{x}+\Delta\mathbf{x})^\top \mathbf{v}_i$ be the mapping of the adversarially perturbed image  on the $i$th eigenvector, respectively. Since the transformation is linear, the change of $p_i$ caused by $\Delta \mathbf{x}$ is $\Delta p_i=\Delta \mathbf{x}^\top \mathbf{v}_i$. The variance of $p_i$ measures the spread of data in the direction of $\mathbf{v}_i$ and hence has $\variance(p_i) = \lambda_i$, while the variance of $p'_i$ has an upper bound of $\lambda_i + \expectation(\Delta p_i^2)$, as shown in \Cref{eq.upperbound},

\begin{align}\label{eq.upperbound}
    \variance(p'_i)  &=  \variance(p_i) + \variance(\Delta p_i) \nonumber \\
    &=  \lambda_i + \expectation(\Delta p_i^2) - \expectation(\Delta p_i)^2 \nonumber \\
    &\leq \lambda_i + \expectation(\Delta p_i^2)
\end{align}
assuming that the adversarial perturbation is independent from the data. Further, the expected deviation of perturbation $\expectation(\Delta p_i^2)$ can be no larger than the norm of the perturbation $\|\Delta\mathbf{x}\|^2$ as shown in \Cref{eq.upperbound_of_derivation}

\begin{alignat}{3}\label{eq.upperbound_of_derivation}
    \expectation(\Delta p_i^2) &= \expectation((\Delta \mathbf{x}^\top \mathbf{v}_i)^2) && \overnumber{1}{\leq} \mathbf{E}(\|\Delta\mathbf{x}\|^2 \|\mathbf{v}_i\|^2) \nonumber \\
    &\overnumber{2}{=}  \mathbf{E}(\|\Delta\mathbf{x}\|^2)  && \overnumber{3}{=} \|\Delta\mathbf{x}\|^2
\end{alignat}
where $\circlednumber{1}$ is due to the Cauchy–Schwarz inequality, $\circlednumber{2}$ holds as $\mathbf{v}_i$ is an eigenvector with  $\|\mathbf{v}_i\|=1$, and $\circlednumber{3}$ holds since for adversarial attacks, the injected perturbation budget $\|\Delta \mathbf{x}\|$ is the same for all images (if the maximum budget is always achieved). By combining \Cref{eq.upperbound,eq.upperbound_of_derivation}, we obtain that $\variance(p'_i)$ can be no larger than $\lambda_i +  \|\Delta\mathbf{x}\|^2$. The empirical analysis in \Cref{fig:tightness} suggests that the actual variance of perturbed projected perturbed images on the least significant eigenvector is much closer to that upper bound than random noises. 
The perturbation budget in the figure varies from 1/255 to 8/255, and for each budget, the result is the average of 1,000 CIFAR10~\cite{cifar} images.

\begin{figure}[!t]
    \centering
    \begin{minipage}[t]{0.58\textwidth}
        \includegraphics[width=\linewidth]{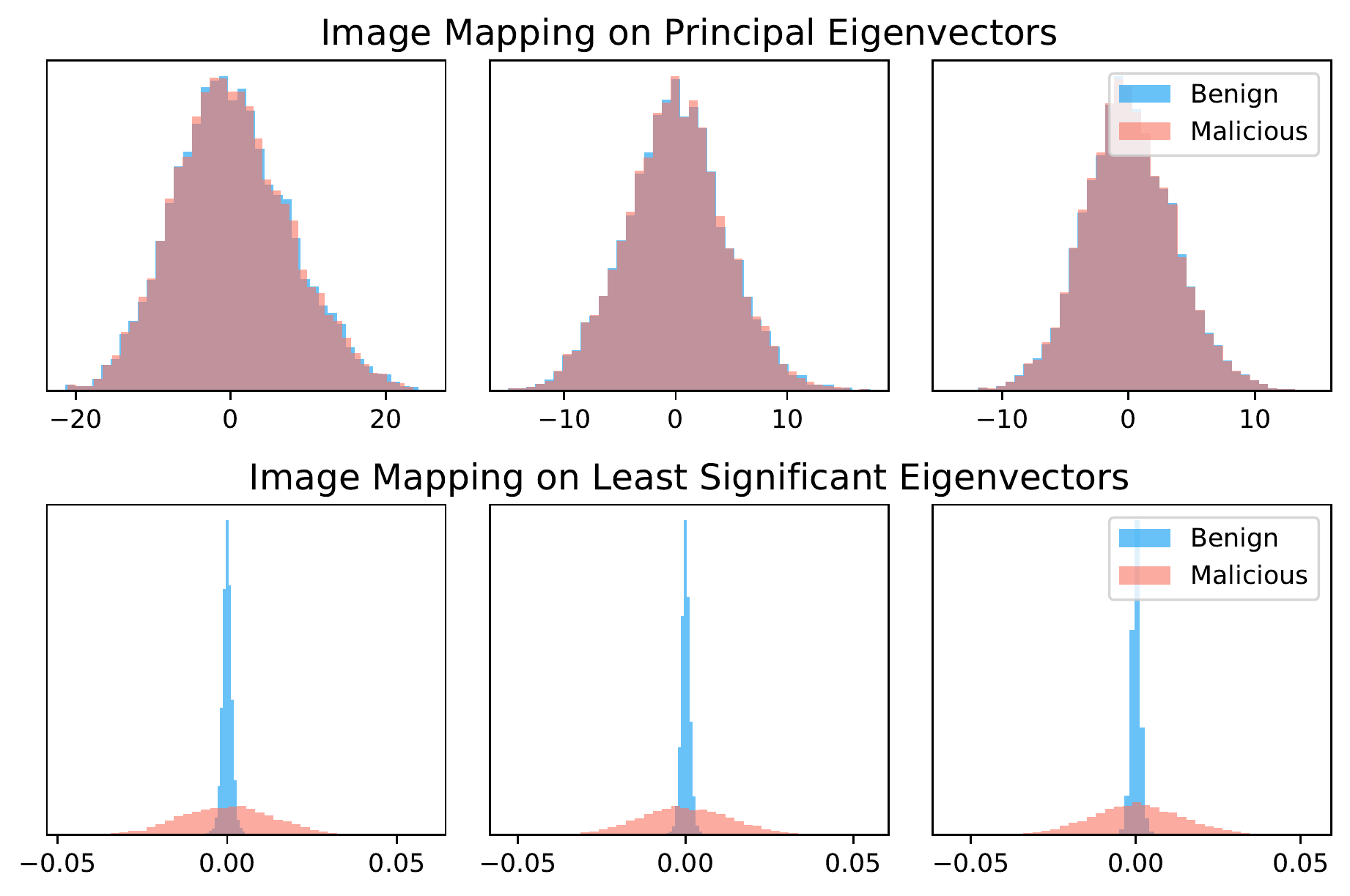}
        \caption{Distribution of images' mapping on three eigenvectors of Principal Components (upper) and \PCAComponent s (bottom) with PGD10 ($\epsilon$=8/255) on CIFAR10.}
        \label{fig:pca_feat_perturb}
    \end{minipage}\hfill
    \begin{minipage}[t]{0.39\textwidth}
        \includegraphics[width=\linewidth]{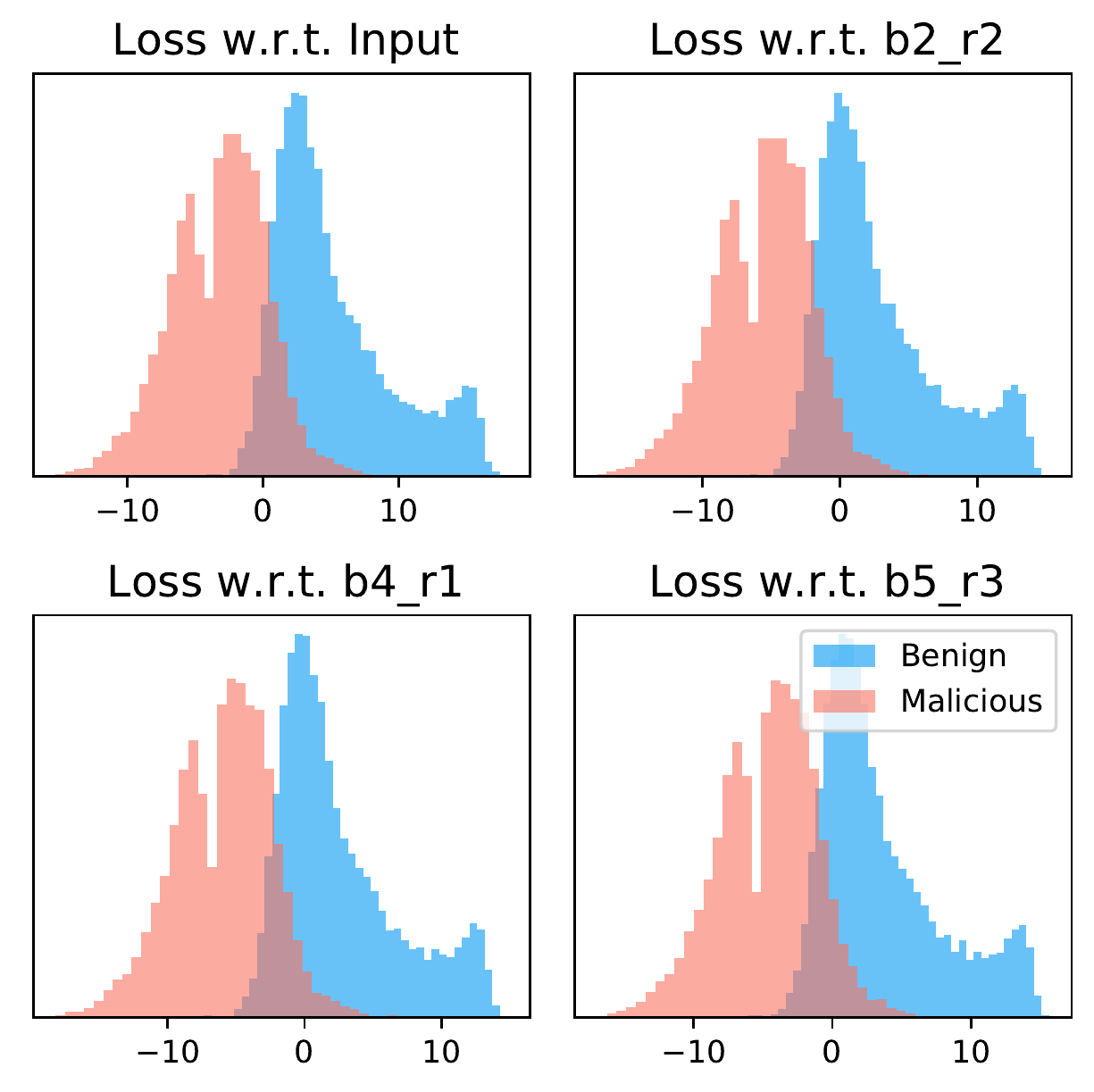}
        \caption{The Hessian modulus distribution of benign and PGD10 ($\epsilon$=8/255) CIFAR10. $bx\_ry$ is block $x$ ReLU $y$ in VGG16.}
        \label{fig:modulus_distribution}
    \end{minipage}

\end{figure}

When the difference between $\variance(p'_i)$ and $\variance(p_i)$ is large, we can easily distinguish benign images from adversarial images by mapping them onto eigenvector $\mathbf{v}_i$. We quantify this difference by the ratio of $\variance(p'_i) / \variance(p_i)$, which has an upper bound  of $1 + \|\Delta \mathbf{x}\| / \lambda_i$, from \Cref{eq.upperbound,eq.upperbound_of_derivation}. 
Since the perturbation budget $\|\mathbf{x}\|$ is predefined before attack, the value of $\lambda_i$ determines the differentiability between $p'_i$ and $p_i$. The smaller the value of $\lambda_i$, the easier to distinguish adversarial from benign images. As a result, mapping data onto the least significant components gives highest distinguishability of attack. \Cref{fig:pca_feat_perturb} visualizes the distribution of projected values for 1,000 adversarial and benign images on major principal components and least significant components. We can see that the distributions for the two types of images are indistinguishable in the major PCA components, but are clearly distinguishable in the least significant components. 

\subsection{\HessianFeat}\label{sec.hessfeat}

When studying model optimization, \cite{jastrzkebski2017three,zhu2018anisotropic} observed that perturbation on model weights can improve generalization, and \cite{wang2020assessing,tsuzuku2020normalized,wang2018identifying} later proved that such improvement happens because the perturbation changes the smoothness of the loss function's landscape, which can be measured by the Hessian matrix of the loss. Motivated by this observation, we hypothesize that the Hessian can be used to characterize the loss landscape and find locations that are exploited by the adversarial perturbations. 
In fact, the adversarial attack creation problem is very similar to the problem of model optimization in the sense that they bear similarity to Lagrange duality.

More formally, model optimization can be expressed by \Cref{eq.model_optim}
\begin{align}\label{eq.model_optim}
    \underset{\mathbf{W}}{\mathrm{minimize}} \quad & L[Y, f(\mathbf{W}, \mathbf{X})] \qquad
     \mathrm{s.t.} \quad \mathbf{X} = \mathbf{D}
\end{align}
while the target of adversarial attack can be written as \Cref{eq.adv_atk} 
\begin{align}
\label{eq.adv_atk}
    \underset{\mathbf{X}}{\mathrm{maximize}} \quad & L[Y, f(\mathbf{W^*}, \mathbf{X})] + \sum_{u_i} u_i \|\mathbf{X}_i - \mathbf{D}_i\|_p \\
    & \mathrm{s.t.} \quad u_i \leq 0, \quad \forall i \in [1, N] \nonumber
\end{align}
where $L$ is the loss function, $\mathbf{W}$ represents the parameters of the model $f$, $\mathbf{D}$ is the training data, $Y$ is corresponding target, and $\mathbf{W^*}$ is the optimized (\ie\ trained) model weights that achieves $\underset{\mathbf{W}}{\mathrm{inf}}(L[Y, f(\mathbf{W}, \mathbf{X})])$. If we regard the Lagrange regularizers as the adversarial perturbation constraints (\ie, $l_\infty$, $l_2$ or $l_0$), \Cref{eq.adv_atk} can be seen as the adversarial attack against the dataset where $L[Y, f(\mathbf{W^*}, \mathbf{X})]$ corresponds to maximizing the prediction error and $\sum_{u_i} u_i \|\mathbf{X}_i - \mathbf{D}_i\|_p$ corresponds to limiting the perturbation budget under $l_p$ constraint.\footnote{We slightly abuse the name of Lagrange regularizer as the norm $\|\cdot\|_p$ is not required in Lagrange duality.} Such correspondence in duality motivates us to measure the statistical deviation of the Hessian matrix to detect adversarial examples. The Hessian we use is the second-order derivative of the loss \wrt the input or the outputs of the intermediate layers, \ie
\begin{equation}
    \mathrm{H} \equiv \frac{\partial^2 L(\mathbf{x})}{\partial^2\mathbf{x}} \quad \text{where} \quad \mathrm{H}[i, j] = \frac{\partial^2 L(\mathbf{x})}{\partial x_i \partial x_j},
\end{equation}
$\mathbf{x}$ is input or the outputs of the intermediate layers of the model (\eg, outputs of ReLU layers), $x_i$ and $x_j$ are the $i$th and $j$th entry (\eg, pixels in image) of the input, respectively. The  size Hessian matrix is proportional to the square of the input dimension, which can cause computational and performance problems for anomaly detection models due to the curse of high dimensionality~\cite{curseofhighdim}. For example, the dimension of the Hessian \wrt the input in the case of CIFAR10 images is $3072 \times 3072 = 9,437,184$. This dimensionality is prohibitive for any existing anomaly detector. 

In order to handle this challenge, we use the modulus of the Hessian as an approximation of the Hessian. Specifically, we use $l1$ norm $\| \mathrm{H} \|_1 = \sum_{i, j} | \mathrm{H}[i, j] |$ of the matrix as we empirically found that the result of using $l1$ and $l2$ norm does not have big difference. The modulus operation reduces the spatial dimension of Hessian matrix to only one scalar number. \Cref{fig:modulus_distribution} shows the Hessian modulus distribution of benign and PGD10 images. The distributions suggest that the modulus of the Hessian can be used to separate benign and adversarial samples. Nevertheless, our final Hessian feature includes multiple dimensions by using the moduli of Hessian matrices for multiple network layers along with the Hessian matrix for the input.



\input{4_GN_matrix}

%% file: 4_GN_matrix.tex
\subsection{Generalized Gauss-Newton Matrix for Approximating Hessian Matrix}
 We compute the Generalized Gauss-Newton matrix~\cite{schraudolph2002fast,martens2010deep,martens2014new} instead to significantly speed up calculating the Hessian. Let $L$ be the loss, $\mathbf{x}$ be the variable to the loss (\eg\ images) and $\mathbf{z}$ be the inputs of penultimate layer (\eg\ the Softmax layer in DNNs). The GGN can be computed as
\begin{equation}
    \mathrm{G} = (\mathrm{J}^{\mathbf{z}}_{\mathbf{x}})^\top \otimes \mathrm{H}^L_{\mathbf{z}} \otimes \mathrm{J}^{\mathbf{z}}_{\mathbf{x}}
\end{equation}
where $\otimes$ is the matrix multiplication, $\mathrm{J}^{\mathbf{z}}_{\mathbf{x}}$ is the Jacobian of the penultimate layer $\textbf{z}$ \wrt\ the input $\mathbf{x}$ and $\mathrm{H}^L_{\mathbf{z}}$ is the Hessian of loss $L$ \wrt\ penultimate layer $\textbf{z}$. Please note that the ground truth label is not required during computation and any choice of label will give the same result since GGN/Hessian modulus only show the curvature of the loss landscape.
Though GGN approximates Hessian well~\cite{schraudolph2002fast,martens2010deep,martens2014new}, it is unclear how good the modulus of GGN approximates the modulus of Hessian. We empirically show the approximation accuracy by randomly picking 1,000 samples from CIFAR10 and computing their Hessian and GGN. \Cref{fig:GGN_Hessian_approximation} shows the matrix modulus of Hessian and GGN while \Cref{fig:GGN_Hessian_time} summarizes the computation time of Hessian and GGN under different image sizes. We notice that GGN is strongly correlated to Hessian, while being much more computationally efficient to calculate. Therefore we use GNN as a substitute for Hessian~\cite{schraudolph2002fast}.

%% file: 5_experiments.tex
\section{Experimental Evaluation}

{{
\subsection{Benchmarks and Baselines}\label{sec.expt_setting}
To evaluate the performance of \ours, we conduct a series of experiments on the CIFAR10~\cite{cifar}, CIFAR100~\cite{cifar}, and SVHN~\cite{svhn} datasets and compare \ours's performance using several anomaly detection methods. As described in Section~\ref{sec:h3ad}, \ours-based anomaly detectors are trained on benign images only. We base our experiments on the VGG16~\cite{vgg} model. There is no specific reason for this choice of model, beyond convenience and (relatively lower) computational requirements, the methodology itself is model-agnostic. All of results are obtained with \textit{Nvidia} 1080Ti GPUs.

\noindent \textbf{Baseline features:} To provide a broad set of comparative baseline features, we compare \ours\ against one naive image feature (PCA), two hand-crafted features (LID~\cite{lid} and Mahalanobis~\cite{mahalanobis}), and one self-trained deep feature (DSVDD~\cite{dsvdd}). We extract 32-dimensional principal components for PCA. Our choice of LID and Mahalanobis is driven by the fact that they do not require supervision to compute features and have less complexity than other methods such as \cite{libre}, which requires the modification of the underlying model followed by finetuning. For both LID and Mahalanobis, we follow the original papers but change the target network to VGG16\cite{vgg} to provide a fair comparison to the \ours\ features.
DSVDD integrates both the feature extractor and anomaly detector. We train the feature extractor for 100 epochs and tune the anomaly detector for 50 epochs, following \cite{dsvdd}.

\noindent \textbf{\ours\ features:} As explained in \ref{sec:h3ad}, \ours\ features consists of two parts. First, we extract a 32-dimensional LSCF feature, as described in \Cref{sec.pca}. We then compute the Hessian feature by evaluating the Hessian of the loss \wrt ~the input and the intermediate features from the ReLU layers to form a 13-dimensional feature, as described in \Cref{sec.hessfeat}. The LSCF and Hessian features are concatenated to create a 45-dimensional \ours\ feature for each image.

\noindent \textbf{Anomaly Detectors:} We train both kernel density estimator (KDE) and One-Class SVM (OCSVM) based anomaly detectors on each set of features. For KDE, we evaluate using Gaussian, Epanechnikov, exponential, linear, and uniform kernels. For OCSVM, we evaluate using RBF, Sigmoid, linear and polynomial kernels. We also conduct a grid search for hyperparameters and report the best performance, the corresponding ablation studies can be found in the Appendix.

\noindent \textbf{Adversarial Attacks:} Each anomaly detector is evaluated across eight standard attacks. For $l_\infty$ attacks with max perturbation 8/255, we use (1) PGD10~\cite{pgd}, (2) FGSM~\cite{fgsm} and (3) BIM~\cite{bim}. For $l_2$ attacks with total perturbation budget of 1, we use (4) DeepFool~\cite{deepfool} and (5)  CW~\cite{cw}. For $l_0$ attacks, we use (6) OnePixel~\cite{onepixel} and (7) SparseFool~\cite{sparsefool} with hyperparameter $lam=3$. For combined attacks with perturbation budget equals to 0.3 under $l_\infty$, we use (8) AutoAttack~\cite{autoattack}. We use the \textit{torchattacks}'~\cite{kim2020torchattacks} implementation for all attacks.

\subsection{Evaluation Results}\label{sec.baseline}
Each anomaly detector is evaluated using the area under receiver operating characteristic curve (ROC AUC) on all adversarial attacks. We report performance on both each attack variant as well as the overall performance across all eight attacks. The results are summarized in \Cref{tab:compare_to_baseline}. Note that, unlike supervised training where the detector is trained and evaluated on a specific attack, here, the same anomaly detector is used to detect \textbf{any} of the eight attacks. 

We observe that, with very few exceptions, across all anomaly detection variants,  \ours-based anomaly detectors demonstrate the best performance. In general, features that represent holistic image features, such as PCA and DSVDD, do not perform well. The subtle and localized adversarial perturbations are likely overwhelmed by these global image features.

\ours\ features, in particular, perform well against both $l_\infty$ attacks and AutoAttack. We find that AutoAttack is easy to detect for all but the PCA-based anomaly detectors. We speculate that the reason for this behavior is that ensemble attacks leave more traces of tampering and are therefore easier to detect. \ours\ features appear to be particularly robust to $l_\infty$ attacks vis-\'a-vis the other approaches. Even on $l_2$ and $l_0$ attacks, \ours\ features perform better than most of the compared features. Across all attacks, \ours\ features achieve almost 95\% AUC on CIFAR10 and SVHN, and almost 90\% AUC on CIFAR100.
}}


\begin{table}[!t]
    \centering
    \scriptsize
    \setlength{\tabcolsep}{2.5pt}
    \begin{tabular}{llccccccccc}
    \toprule
       &  & \multicolumn{3}{|c|}{$l_\infty$ Attacks} & \multicolumn{2}{c|}{$l_2$ Attacks} & \multicolumn{2}{c|}{$l_0$ Attacks} & \multicolumn{1}{c|}{Combined} & \multirow{2}{*}{Overall} \\
       \cmidrule(lr){3-5} \cmidrule(lr){6-7} \cmidrule(lr){8-9} \cmidrule(lr){10-10}
       Dataset  & Method & PGD10 & FGSM & BIM & DeepFool & CW & SparseFool & OnePixel & AutoAttack & \\
       \cmidrule(lr){1-1} \cmidrule(lr){2-2} \cmidrule(lr){3-3} \cmidrule(lr){4-4}
       \cmidrule(lr){5-5} \cmidrule(lr){6-6}
       \cmidrule(lr){7-7} \cmidrule(lr){8-8}
       \cmidrule(lr){9-9} \cmidrule(lr){10-10} \cmidrule(lr){11-11} 
       \multirow{9}{*}{CIFAR10} & PCA+OCSVM & 0.497 & 0.498 & 0.497 & 0.500 & 0.500 & 0.109 & 0.497 & 0.293 & 0.424 \\
       & PCA+KDE & 0.501 & 0.498 & 0.500  & 0.501 & 0.500 & 0.502 & 0.500 & 0.501 & 0.500 \\
       & DSVDD~\cite{dsvdd} & 0.569 & 0.614 & 0.566 & 0.505 & 0.507 & 0.901 & 0.505 & 0.958 & 0.641 \\
       & LID+OCSVM & 0.551 & 0.596 & 0.575 & 0.583 & 0.585 & 0.914 & 0.559 & 0.968 & 0.666\\
       & LID~\cite{lid}+KDE & 0.610 & 0.702 & 0.639 & 0.654 & 0.656 & 0.924 & 0.615 & 0.971 & 0.721 \\
       & Mah.+OCSVM & 0.880 & 0.787 & 0.898 & \textbf{0.852} & 0.837 & 0.963 & 0.668 & \textbf{0.989} & 0.859 \\
       & Mah.~\cite{mahalanobis}+KDE & 0.896 & 0.887 & 0.893 & 0.603 & 0.587 & 0.899 & 0.578 & 0.966 & 0.789 \\
       \cmidrule(lr){2-2} \cmidrule(lr){3-11}
       & \ours+OCSVM (Ours) & \underline{0.999} & \textbf{0.999} & \underline{0.999} & 0.841 & \underline{0.941} & \underline{0.985} & \underline{0.821} & 0.988 & \underline{0.947} \\
       & \ours+KDE (Ours) & \textbf{1.000} & \textbf{0.999} & \textbf{1.000} & \underline{0.846} & \textbf{0.943} & \textbf{0.986} & \textbf{0.825} & \textbf{0.989} & \textbf{0.949} \\
       \cmidrule(lr){1-11}
       \multirow{9}{*}{CIFAR100} & PCA+OCSVM & 0.497 & 0.497 & 0.497 & 0.500 & 0.500 & 0.221 & 0.497 & 0.353 & 0.445 \\
       & PCA+KDE & 0.498 & 0.501 & 0.499 & 0.500 & 0.500 & 0.501 & 0.500 & 0.497 & 0.500\\
       & DSVDD & 0.568 & 0.629 & 0.564 & 0.502 & 0.504 & 0.777 & 0.501 & 0.852 & 0.612 \\
       & LID+OCSVM & 0.570 & 0.579 & 0.581 & 0.504 & 0.501 & 0.758 & 0.520 & 0.845 & 0.607\\
       & LID+KDE & 0.642 & 0.655 & 0.654 & 0.511 & 0.515 & 0.768 & 0.549 & 0.849 & 0.643 \\
       & Mah.+OCSVM & 0.708 & 0.719 & 0.709 & \textbf{0.816} & 0.811 & 0.772 & \textbf{0.883} & \textbf{0.916} & 0.792\\
       & Mah.+KDE & 0.845 & 0.926 & 0.848 & 0.535 & 0.541 & 0.760 & 0.530 & 0.798 & 0.723 \\
       \cmidrule(lr){2-2} \cmidrule(lr){3-11}
       & \ours+OCSVM (Ours) & \textbf{0.999} & \underline{0.999} & \textbf{0.998} & 0.728 & \underline{0.814} & \underline{0.898} & 0.819 & 0.906 & \underline{0.895} \\
       & \ours+KDE (Ours) & \textbf{0.999} & \textbf{1.000} & \textbf{0.998} & \underline{0.733} & \textbf{0.816} & \textbf{0.901} & \underline{0.820} & \underline{0.908} & \textbf{0.897} \\
       \cmidrule(lr){1-11}
       \multirow{9}{*}{SVHN} & PCA+OCSVM & 0.499 & 0.501 & 0.499 & 0.500 & 0.500 & 0.242 & 0.497 & 0.342 & 0.448 \\
       & PCA+KDE & 0.500 & 0.499 & 0.499 & 0.499 & 0.499 & 0.501 & 0.500 & 0.495 & 0.499 \\
       & DSVDD & 0.717 & 0.812 & 0.714 & 0.524 & 0.527 & 0.911 & 0.521 & 0.981 & 0.713 \\
       & LID+OCSVM & 0.680 & 0.640 & 0.693 & 0.654 & 0.680 & 0.927 & 0.525 & 0.984 & 0.723\\
       & LID+KDE & 0.761 & 0.747 & 0.772 & 0.726 & 0.749 & 0.938 & 0.560 & 0.986 & 0.780 \\
       & Mah.+OCSVM & 0.747 & 0.699 & 0.766 & \textbf{0.917} & 0.941 & 0.966 & 0.663 & \textbf{0.994} & 0.837 \\
       & Mah.+KDE & 0.833 & 0.748 & 0.848 & 0.904 & 0.914 & 0.909 & 0.638 & 0.971 & 0.846 \\
       \cmidrule(lr){2-2} \cmidrule(lr){3-11}
       & \ours+OCSVM (Ours) & \textbf{1.000} & \textbf{1.000} & \textbf{1.000} & 0.868 & \underline{0.975} & \underline{0.992} & \textbf{0.954} & 0.993 & \underline{0.934} \\
       & \ours+KDE (Ours) & \textbf{1.000} & \textbf{1.000} & \textbf{1.000} & \textbf{0.917} & \textbf{0.979} & \textbf{0.994} & \underline{0.946} & \textbf{0.994} & \textbf{0.946}\\
       \bottomrule
    \end{tabular}
    \caption{The ROC AUC performance on detecting eight adversarial attacks. Best performance is reported in \textbf{bold} and second best with \underline{underline}.}
    \label{tab:compare_to_baseline}

\end{table}

\subsection{Cross-model Adversarial Detection}\label{sec.cross_model}
Adversarial examples generated by one model are known to be transferrable in that they can deceive a trained model with a different architecture~\cite{transferattack}. In such cases, the defender attempts to detect adversarial images generated by an unknown model. To evaluate this scenario, we generate adversarial images with a ResNet18~\cite{resnet} model and the defender's task is to protect a VGG16~\cite{vgg} model. For cross-model adversarial detection with LID~\cite{lid} and Mahalanobis~\cite{mahalanobis}, we find that the baseline anomaly detectors perform quite poorly. To provide a stronger comparison, we instead compare against the LID and Mahalanobis supervised models. (Note that supervised models are trained on adversarial images of VGG16 but evaluated on adversarial images of ResNet18.) The supervised model is a binary classifier consisting of four fully connected layers with output dimensions of 64, 32, 8, and 1. ReLU layers and batch normalization layers~\cite{batchnorm} are attached after the first three fully connected layers, and Sigmoid layer after the last one. We optimize this model with SGD~\cite{sgd}, with learning rate = 0.001, for 100 epochs using binary cross-entropy loss. For \ours\ however, we use the same anomaly detection models as in \Cref{sec.expt_setting}. As shown in \Cref{tab:cross_model}, across all datasets and attacks, \ours\ based anomaly detectors significantly outperforms the supervised LID and Mahalanobis feature based models. Only on $l_{2}$ DeepFool attack, Mahlanobis-based supervised model slightly outperforms the \ours-based anomaly detector. To reiterate, a \ours\ based anomaly detector, trained on benign images only, outperforms supervised models, trained on LID or Mahalanobis features, on the cross-model adversarial detection task.

\begin{table}[!t]
    \centering
    \scriptsize
    \setlength{\tabcolsep}{2.5pt}
    \begin{tabular}{llccccccccc}
    \toprule
       &  & \multicolumn{3}{|c|}{$l_\infty$ Attacks} & \multicolumn{2}{c|}{$l_2$ Attacks} & \multicolumn{2}{c|}{$l_0$ Attacks} & \multicolumn{1}{c|}{Combined} & \multirow{2}{*}{Overall} \\
       \cmidrule(lr){3-5} \cmidrule(lr){6-7} \cmidrule(lr){8-9} \cmidrule(lr){10-10}
       Dataset  & Method & PGD10 & FGSM & BIM & DeepFool & CW & SparseFool & OnePixel & AutoAttack & \\
       \cmidrule(lr){1-1} \cmidrule(lr){2-2} \cmidrule(lr){3-3} \cmidrule(lr){4-4}
       \cmidrule(lr){5-5} \cmidrule(lr){6-6}
       \cmidrule(lr){7-7} \cmidrule(lr){8-8}
       \cmidrule(lr){9-9} \cmidrule(lr){10-10} \cmidrule(lr){11-11} 
       \multirow{4}{*}{CIFAR10} 
       & LID (Binary Classification) & 0.594 & 0.901 & 0.588 & \underline{0.617} & 0.652 & 0.830 & 0.677 & 0.876 & 0.717 \\
       & Mah. (Binary Classification) &  0.702 & 0.991 & 0.704 & \textbf{0.628} & 0.658 & 0.838 & 0.674 & 0.740 & 0.742 \\
       \cmidrule(lr){2-2} \cmidrule(lr){3-11}
       & \ours+OCSVM (Ours) & \textbf{1.000} & \textbf{1.000} & \textbf{0.999} & 0.589 & \underline{0.880} & \underline{0.969} & \textbf{0.882} & \textbf{0.988} & \underline{0.913} \\
       & \ours+KDE (Ours) & \textbf{1.000} & \textbf{1.000} & \textbf{0.999} & 0.590 & \textbf{0.883} & \textbf{0.970} & \underline{0.881} & \textbf{0.988} & \textbf{0.914} \\
       \cmidrule{1-11}
       \multirow{4}{*}{CIFAR100} 
       & LID (Binary Classification) & 0.650 & 0.831 & 0.636& 0.499 & 0.504 & 0.810 & 0.596 & 0.842 & 0.671 \\
       & Mah. (Binary Classification) & 0.737 & 0.985 & 0.713 & \textbf{0.531} & 0.556 & 0.839 & 0.662 & 0.752 & 0.722  \\
       \cmidrule(lr){2-2} \cmidrule(lr){3-11}
       & \ours+OCSVM (Ours) & \textbf{0.999} & \textbf{0.999} & \textbf{0.998} & 0.527 & \underline{0.762} & \underline{0.901} & \underline{0.814} & \underline{0.917} & \underline{0.861} \\
       & \ours+KDE (Ours) & \textbf{0.999} & \textbf{0.999} & \textbf{0.998} & \underline{0.530} & \textbf{0.765} & \textbf{0.905} &\textbf{ 0.815} & \textbf{0.919} & \textbf{0.866} \\
       \cmidrule{1-11}
       \multirow{4}{*}{SVHN} 
       & LID (Binary Classification) & 0.776 & 0.777 & 0.788 & 0.593 & 0.609 & 0.931 & 0.583 & 0.890 & 0.743 \\
       & Mah. (Binary Classification) &  0.797 & 0.847 & 0.808 &\textbf{0.647} & 0.659 & 0.942 & 0.651 & 0.801 & 0.769 \\
       \cmidrule(lr){2-2} \cmidrule(lr){3-11}
       & \ours+OCSVM (Ours) & \textbf{1.000} & \textbf{1.000} & \textbf{1.000} & \underline{0.605} & \underline{0.898} & \textbf{0.993} & \underline{0.928} & \underline{0.994} & \underline{0.927} \\
       & \ours+KDE (Ours) & \textbf{1.000} & \textbf{1.000} & \textbf{1.000} & 0.602 & \textbf{0.901} & \underline{0.992} & \textbf{0.930} & \textbf{0.995} & \textbf{0.928} \\
       \bottomrule
    \end{tabular}
    \caption{The ROC AUC performance on detecting cross model adversarial attacks. Best performance is reported in \textbf{bold} and second best with \underline{underline}.}
    \label{tab:cross_model}

\end{table}

\subsection{Sensitivity and Ablation Studies}\label{sec.ablation}
To further understand the properties of the \ours\ features, we conduct experiments on CIFAR10 to evaluate (i) effectiveness of \pcafeat\ and \hessianfeat, (iii) performance gap between anomaly detection and binary classification, (iii) method sensitivity to the anomaly detectors, and (iv) method robustness when distinguish benign noisy images and adversarial images. The result of (iii) and (iv) are provided in the Appendix due to page limitation.

\noindent \textbf{Effectiveness of \PCAFeat\ And \HessianFeat\ components of \ours: }
To compare the effectiveness of \PCAFeat\ and \HessianFeat\ we ablate on the number of feature components. Specifically, for \pcafeat, we use 0, 4, 16, 32, 64-dimensional feature variants. For \hessianfeat, we use 0, 1 (only input), 5 (from input to $b2\_r2$), 9 (from input to $b4\_r1$) and 13-dimensional (from input to $b5\_r3$) features. 
When one feature size (\pcafeat ~or \hessianfeat) is changed, we use the best number of feature components for the other feature. Results are detailed in the \Cref{tab:ablation_dim}. Both features show improved performance as the number of feature components increases. We observe that \pcafeat\ and \hessianfeat\ are complementary in that the largest performance gains are obtained when \pcafeat\ and \hessianfeat\ are concatenated. For \pcafeat, performance plateaus at 32 dimensions and does not increase with doubling the number of dimensions. Based on this ablation study, we choice 13-dimensional \hessianfeat\ and 32-dimensional \pcafeat\ in the experiments of the remaining paper.

\begin{table}[!htbp]
    \centering
    \scriptsize

    \begin{minipage}[t]{0.48\textwidth}
    \centering
        \begin{tabular}{r|cc}
        \toprule
          \textbf{\hessianfeat\ Dimension}  & \textbf{ROC AUC} & \textbf{Improve}  \\
          \cmidrule(lr){1-1} \cmidrule(lr){2-2} \cmidrule(lr){3-3} 
          0 & 0.885 & - \\
          1 & 0.936 & +0.051 \\
          5 & 0.946 & +0.010\\
          9 & 0.948 & +0.002\\
          13 & 0.949 & +0.001 \\
        \bottomrule
        \end{tabular}
    \end{minipage}\hfill
    \centering
    \begin{minipage}[t]{0.48\textwidth}
    \centering
        \begin{tabular}{r|cc}
        \toprule
        \textbf{\pcafeat\ Dimension} & \textbf{ROC AUC} & \textbf{Improve} \\
        \cmidrule(lr){1-1} \cmidrule(lr){2-2}  \cmidrule(lr){3-3} 
        0 & 0.860 & - \\
        4 & 0.923 & +0.063 \\
        16 & 0.939 & +0.016\\
        32 & 0.949 & +0.010\\
        64 & 0.949 & +0.000\\
        \bottomrule
        \end{tabular}
        
    \end{minipage}
    \caption{The effectiveness of different dimensional \HessianFeat\ (left) and \PCAFeat\ (right). The performance is shown in ROC AUC over all attacks. Dimension=0 implies the feature is not used. The right column shows the incremental performance improvement over the prior row.}
    \label{tab:ablation_dim}

\end{table}

\begin{figure}[!t]
    \centering
    \includegraphics[width=\linewidth]{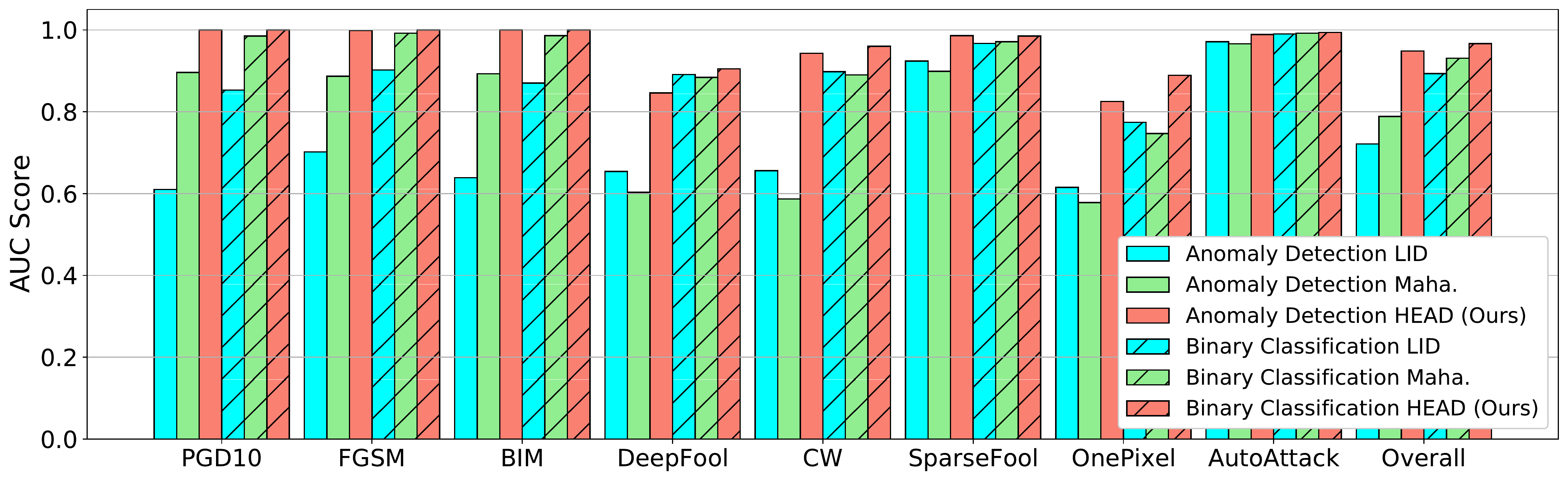}
    \caption{The performance of adversarial detection using anomaly detection and binary classification on CIFAR10. The results of anomaly detection and binary classification are shown in pure color bars and shadowed bars, respectively. The overall performance for binary classification is the average performance of eight attacks.}
    \label{fig:compare_to_supervision}

\end{figure}

\noindent \textbf{Binary benign/attack classification vs. anomaly detection:} 
Anomaly detection, in general, does not require examples of the outliers, \ie\ the adversarial images in this study. An interesting question is what, if any, performance improvement could be gained by incorporating knowledge of the adversarial examples? To answer this question, we use a binary benign/attack classifier to provide an upper bound on the performance, where we train neural networks on benign and adversarial images as inputs with image class (benign or adversarial) as the output. The binary classifier has the same architecture as the previously described LID and Mahalanobis  models in \Cref{sec.cross_model}. \Cref{fig:compare_to_supervision} compares supervised training with anomaly detection using three different input features: LID, Mahanolobis, and \ours. Across all input features, we find that supervised training provides better performance over the corresponding anomaly detector.  However, anomaly detection using \ours\ features \emph{of only benign samples} show a much smaller performance gap with supervised training, and performance is quite comparable, reinforcing the suitability of \ours\ features for adversarial image detection, since knowledge of potential attacks and the ability to generate samples from these attacks is not practical.


%% file: 6_conclusion.tex
\section{Conclusion}\label{sec:conclusion}

We frame adversarial detection as an anomaly detection problem to better reflect the challenge of detecting adversarial examples in real life. We propose \Ours, which measures the statistical deviation caused by adversarial perturbation on two complementary features: \pcafeat, which captures the deviation of adversarial images from the benign data, and \hessianfeat, which reflects the deformation of the model's loss landscape at adversarialy perturbed images.
We provide the theoretical rationale behind using \pcafeat\ and \hessianfeat. We propose using the Generalized Gauss-Newton as a very efficient and faithful approximation to the Hessian matrix in \hessianfeat. 
Empirical results prove the effectiveness of \ours\ and show that comparable performance to binary classification based adversarial detection can be achieved with anomaly detection. Our method does not use any outlier examples upon training anomaly detection, which could be a limitation in cases where outlier examples are easy to obtain. We defer the study of this case to our future research.

%% file: appendix.tex
\section{Additional Ablation Studies}

\begin{figure}[!htbp]
    \centering
    \begin{minipage}[t]{0.44\textwidth}
    \centering
        \includegraphics[width=\linewidth]{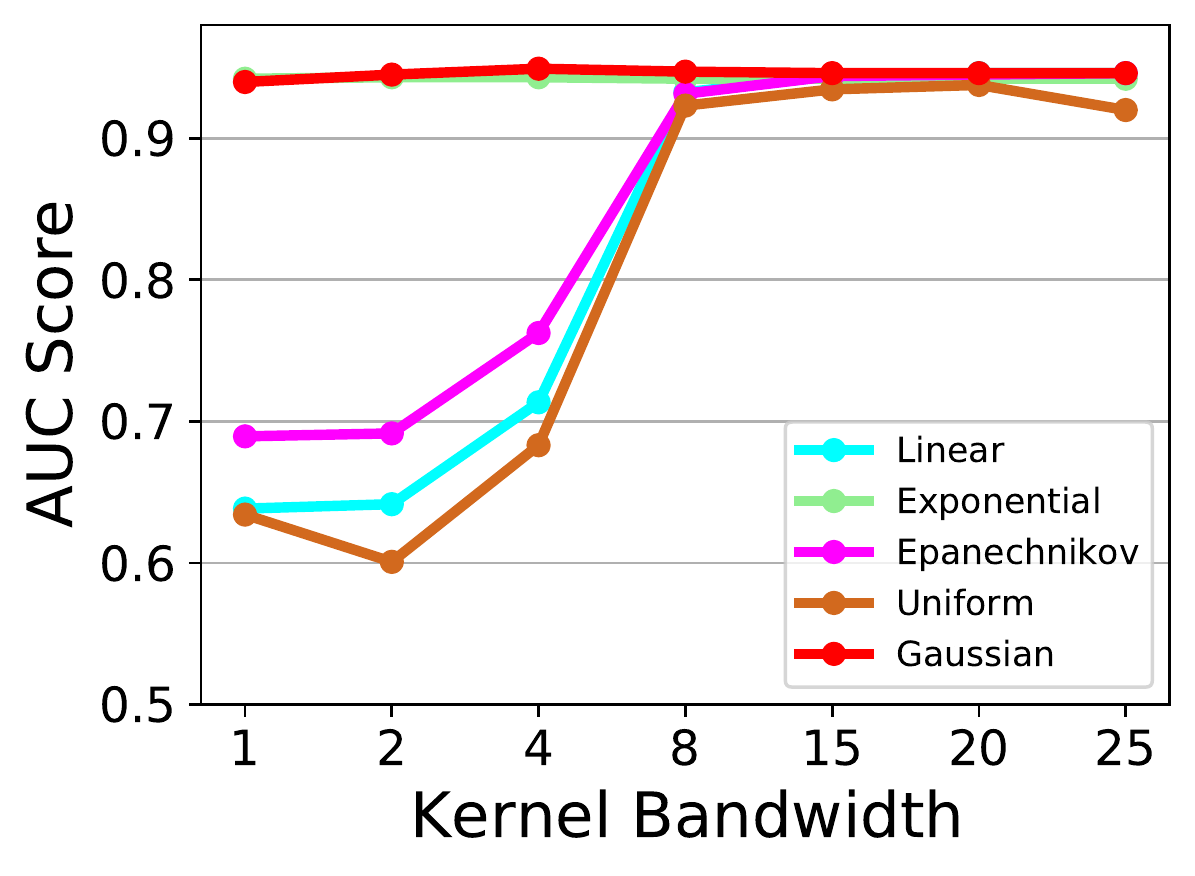}
        \caption{Ablation study of using different KDE kernels and kernel bandwidth. }
        \label{fig:kde_ablation}
    \end{minipage}\hfill
    \begin{minipage}[t]{0.44\textwidth}
    \centering
        \includegraphics[width=\linewidth]{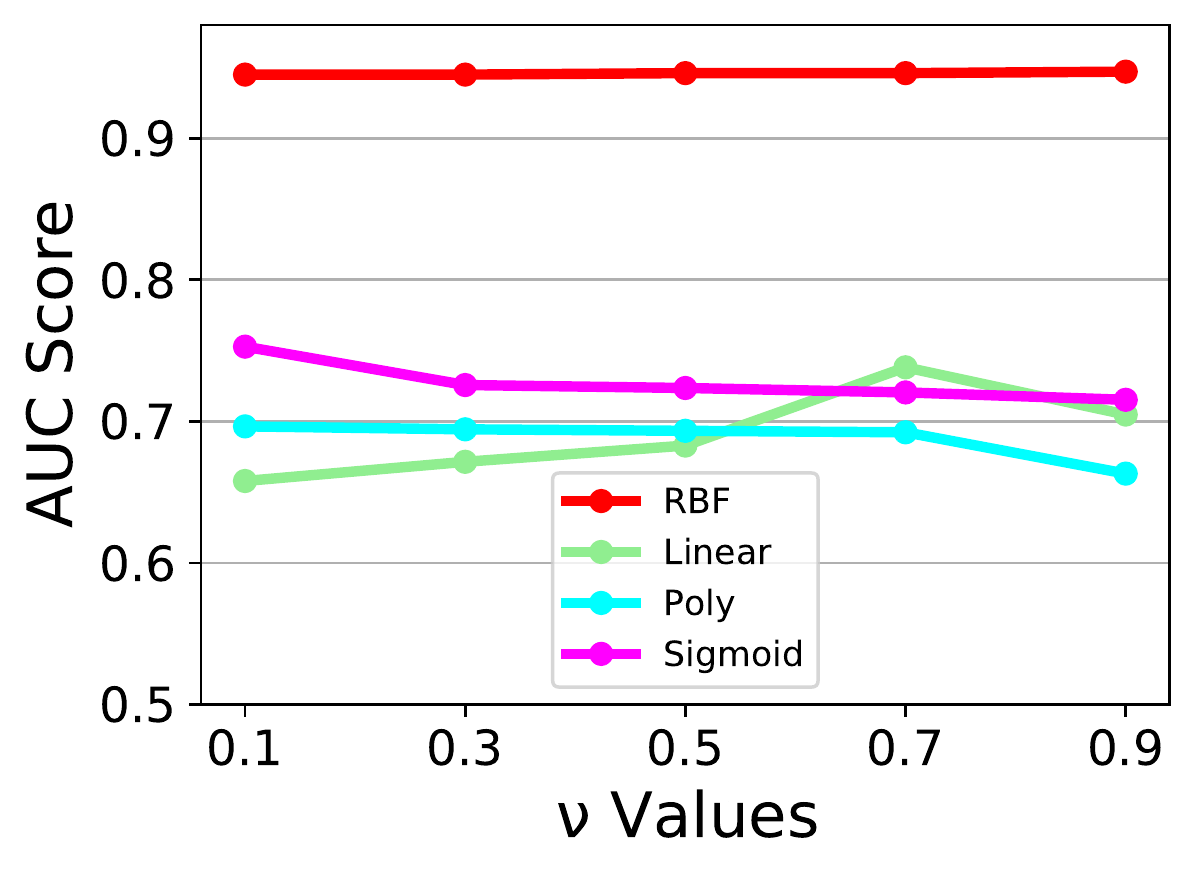}
        \caption{Ablation study of using different OCSVM kernels and $\nu$ values. }
        \label{fig:ocsvm_ablation}
    \end{minipage}
\end{figure}

\noindent \textbf{Sensitivity to anomaly detector parameters:} KDE requires a choice of kernel and bandwidth, and OCSVM requires a selection of kernel and $\nu$ value. We evaluate KDE using Gaussian, Epanechnikov, exponential, linear, and uniform kernels with bandwidth values from 1 to 25. \Cref{fig:kde_ablation} shows the overall AUCs for these parameter values. The results indicate the choice of the kernel is not critical, since all kernels achieve similar performance with an appropriate bandwidth choice. For OCSVM, we evaluate using RBF, Sigmoid, linear and polynomial kernels with $\nu$ values from 0.1 to 0.9. Results are shown in \Cref{fig:ocsvm_ablation}. Unlike KDE, OCSVM is sensitive to the choice of kernel, with the RBF kernel significantly outperforming all other  kernels. That said, with an appropriate choice of hyperparameters,  \ours-based detector performance is insensitive to the choice of anomaly detector. 


\begin{table}[!h]
\centering
    \begin{tabular}{c|cc|cc}
    \toprule
        \textbf{Noise Type} & \multicolumn{2}{c|}{\textbf{Gaussian}} & \multicolumn{2}{c}{\textbf{Uniform}} \\
        \cmidrule(lr){1-1} \cmidrule(lr){2-3} \cmidrule(lr){4-5}
        Noise Level & AUC & Drop & AUC & Drop \\
        \cmidrule(lr){1-1} \cmidrule(lr){2-2} \cmidrule(lr){3-3} \cmidrule(lr){4-4} \cmidrule(lr){5-5}
        0 & 0.949 & - & 0.949 & - \\
        1/255 & 0.929 & -0.020 & 0.934 & -0.015 \\
        2/255 & 0.910 & -0.019 & 0.920 & -0.014 \\
        4/255 & 0.886 & -0.024 & 0.900 & -0.020 \\
        \hspace{-1px}\cellcolor{gray!25}8/255 & \cellcolor{gray!25}0.867 & \cellcolor{gray!25}-0.019 & \cellcolor{gray!25}0.880 & \cellcolor{gray!25}-0.020 \\
        16/255 & 0.834 & -0.033 & 0.856 & -0.024 \\
        32/255 & 0.784 & -0.050 & 0.813 & -0.043 \\
    \bottomrule
    \end{tabular}
    \caption{Performance of adversarial anomaly detector on distinguishing noisy benign images and adversarial images.}
    \label{tab:against_noise}
\end{table}

\noindent \textbf{Robustness To Harmless Random Noise:} 
While random noise can be viewed as a perturbation to clean images, they do not generally result in wrong predictions except at high noise levels.  A good adversarial anomaly detector should be able to distinguish noisy benign images from adversarial images. To evaluate this behavior we train anomaly detectors on benign images (without noise) and test on noisy benign images and adversarial images. As additive noise, we use either zero-mean Gaussian noise with standard deviation set to a specified noise level, or zero-mean uniform noise with maximum value equal to a specified noise level. \Cref{tab:against_noise} details overall performance under six different noise levels using the KDE detector. The gray band in the table represents the noise level equivalent to the perturbation budget used in the adversarial attacks. We observe that when noise levels are low, the performance of the detectors does not drop significantly, and remains higher than 85\% AUC. Even when the noise level is double that of the adversarial perturbation budget (\ie, noise level=16/255), the performance is still above 80\% AUC. In general, \ours-based anomaly detectors appear to be robust to random noise no larger than perturbation budgets, while experiencing larger performance drop under strong noise (\eg, noise level=32/255).

